# Fitness Dependent Optimizer: Inspired by the Bee Swarming Reproductive Process

**Jaza M. Abdullah[1], Tarik A. Rashid[2] (IEEE Member)**

[1] Jaza Mahmood Abdullah, Information Technology, College of Commerce, The University of Sulaimani, Sulaymaniyah, Iraq. jaza.abdullah@univsul.edu.iq

[2] Tarik A. Rashid, Computer Science and Engineering, University of Kurdistan Hewler, (UKH), Erbil, KRG, Iraq. tarik.ahmed@ukh.edu.krd

**ABSTRACT** In this paper, a novel swarm intelligent algorithm is proposed, known as the fitness dependent optimizer (FDO). The bee swarming reproductive process and their collective decision-making have inspired this algorithm; it has no algorithmic connection with the honey bee algorithm or the artificial bee colony algorithm. It is worth mentioning that FDO is considered a particle swarm optimization (PSO)-based algorithm that updates the search agent position by adding velocity (pace). However, FDO calculates velocity differently; it uses the problem fitness function value to produce weights, and these weights guide the search agents during both the exploration and exploitation phases. Throughout the paper, the FDO algorithm is presented, and the motivation behind the idea is explained. Moreover, FDO is tested on a group of 19 classical benchmark test functions, and the results are compared with three well-known algorithms: PSO, the genetic algorithm (GA), and the dragonfly algorithm (DA), additionally, FDO is tested on IEEE Congress of Evolutionary Computation Benchmark Test Functions (CEC-C06, 2019 Competition) [1]. The results are compared with three modern algorithms: (DA), the whale optimization algorithm (WOA), and the salp swarm algorithm (SSA). The FDO results show better performance in most cases and comparative results in other cases. Furthermore, the results are statistically tested with the Wilcoxon rank-sum test to show the significance of the results. Likewise, FDO stability in both the exploration and exploitation phases is verified and performance-proofed using different standard measurements. Finally, FDO is applied to real-world applications as evidence of its feasibility.

**Index Terms** Optimization, Swarm Intelligence, Evolutionary Computation, Metaheuristic Algorithms, Fitness Dependent Optimizer, FDO.

## I. INTRODUCTION

From the time when computers were invented, searching for the unknown and looking for the best solution were points of focus. As early as 1945, Alan Turing used a type of search algorithm for breaking German Enigma ciphers during World War II [2]. To date, hundreds of types of algorithms have been developed for various purposes, including optimization problems. Optimization algorithms are used to find suitable solutions for a problem. There might be many different solutions for a single problem, but the optimum solution is preferable. Usually, optimization problems are nonlinear with a complex landscape. Generally, optimization algorithms can be classified into traditional and evolutionary algorithms. Traditional algorithms include gradient-based algorithms and quadratic programming. Evolutionary algorithms include heuristic or metaheuristic algorithms and many hybrid techniques.

Traditional algorithms are efficient in their work; however, several facts can be discussed about them. They are mostly deterministic; for example, a given input will always obtain the same output (except hill climbing when using random restart). Moreover, they perform local searches, which is why there is no guarantee that global optimality will be reached for most of the optimization problems. Consequently, they have limited diversity in the obtained solutions. Additionally, they use some information about the problems, and therefore, they tend to be problem-specific. Furthermore, these traditional algorithms cannot effectively solve multimodal problems because they do not work on highly nonlinear problems.

Evolutionary algorithms could be the correct answer to previous limitations as they have stochastic behaviors. They come in two forms: heuristic and meta-heuristic algorithms. Heuristic algorithms search for a solution by trial and error; they hope that a quality solution will be found in a reasonable amount of time. Similarly, they tend to use specific randomization mechanisms and local searches in various ways. More studies and developments have been conducted on heuristic algorithms to make what is known as metaheuristic algorithms. Metaheuristic algorithms have better performance than heuristics algorithms, which is why the "meta" prefix was added, which means "higher" or "beyond". However, researchers currently use these two terms (heuristic and metaheuristic) interchangeably, as there is little difference in their definitions [3] [4] [5].

The complexity of real-world problems that exist around us makes it impossible to search every possible solution simply because of time, space, and cost considerations. As a result, low cost, fast, and more intelligent mechanisms are required. Therefore, researchers have studied the behaviors of animals and natural phenomena to understand how they solve their problems. For example, how ants find their path, how a group of fish, birds or flies avoid the enemy or hunt their prey, and how gravity works. Thus, these algorithms, which are inspired by nature, are known as nature-inspired algorithms.

Development in nature-inspired metaheuristic algorithms began in the 1960s at the University of Michigan. John Holland and his colleagues published their genetic algorithm (GA) book in 1960 and republished it in 1970 and 1983 [6]. An algorithm that is inspired by the annealing process of metal, known as simulated annealing (SA), was developed by S. Kirkpatrick, C. D. Gellar, and M. P. Vecchi, [7]. Nevertheless, in the past two decades, this field has witnessed many major signs of progress. For instance, particle swarm optimization, which was proposed by James Kennedy and Russel C. Eberhart, has been used for many real-world applications [8]. PSO was inspired by the swarm intelligence of fish and birds while the authors were studying a flock of birds. They found that they could apply these behaviors to optimization problems; later, PSO became a base algorithm for other algorithms, including our algorithm. R. Storn and K. Price developed differential evolution (DE) in 1997. It is a vector-based algorithm that outperforms GA in many applications [9]. After that, in 2001, Zong WooGeem et al. developed the harmony search (HS), which was applied in many optimization problems such as transport modeling and water distribution [10]. Then, in 2004, C. Tovey and S. Nakrani developed the honey bee algorithm. They used it for Internet hosting center optimization [11]. This was followed by the development of a novel bee algorithm proposed by D. T. Pham et al. [12], and one year later, D. Karaboga et al. created the artificial bee colony (ABC) algorithm in 2005. In 2009, Xin-She Yang developed the firefly algorithm (FA) [13]; and then, the cuckoo search (CS) algorithm was proposed by the same author [14]. Additionally, Xin-She Yang proposed a bat-inspired algorithm in 2010 [15]. Then, in 2015, Mirjalili A. S. proposed the dragonfly algorithm (DA) [16], which is a PSO-based algorithm inspired by the dragonfly swarm behavior of attraction to food and distraction by the enemy, then the whale optimization algorithm (WOA) in 2016 [17], and the salp swarm algorithm (SSA) in 2017 were proposed by the same author [18]. Two new variants of the ABC are proposed by Laizhong et al, the authors showed that they managed to enhance the exploitations of the novel ABC algorithm, as it is well known that the novel ABC has a good exploration ability, however, it suffers from slow exploitations. In their first work, they employed an adaptive method for the population size (AMPS) [19]. In the second paper, they proposed a ranking-based adaptive ABC algorithm (ARABC) [20], the attention on both works was to improve exploitations ability of the novel ABC. Nonetheless, two more improvements were suggested on the novel ABC in 2018, firstly, by proposing the distance-fitness-based neighbor search mechanism (DFnABC), which is a new variant of the ABC [21], and secondly, by proposing the dual-population framework (DPF), again to enhance ABC convergence speed [22]. Additionally, in 2018, a new algorithm which inspired by vapour-liquid equilibrium (VLE) was proposed by Enrique M. Cortés-Toro and his colleagues, the authors claim that their algorithm can solve highly nonlinear optimization problems in continuous domains [23].

Various research has been conducted in the field of nature-inspired metaheuristic algorithms; additionally, many efficient algorithms have been proposed in the literature. Alternatively, there is always room for new algorithms, as long as the proposed algorithm provides better or comparative performances, as explained by David H. Wolpert and William G. Macready in their work titled "No Free Lunch Theorems for Optimization" in 1997. Thus, there is no single global algorithm that can provide the optimum solution for every optimization problem. For example, if algorithm "A" works better than algorithm "B" on optimization problem X, then there is a high chance that there is an optimization problem Y, that works better on algorithm "B" than on algorithm "A" [24]. For these reasons, a new algorithm (FDO) is proposed in this paper. This algorithm is inspired by the swarming behavior of bees during the reproductive process when they search for new hives. This algorithm has nothing in common with the ABC algorithm (except both algorithms are inspired by bee behavior, and both are nature-inspired meta-heuristic algorithms).

The major contributions of this paper are summarized as follows:

1- A new novel swarm intelligent algorithm is proposed, which is using certain characteristics of the bee swarms. For example, it uses a fitness function for generating suitable weights that help the algorithm in both exploration and exploitation phases, as it provides fast convergence towards global optimality with respect to fair coverage of the search landscape.
2- One more unique feature of FDO is that it stores the past search agent pace (velocity) for potential reuse in future steps (more on this is discussed in section IV).
3- FDO can be considered a PSO-based algorithm since it uses a similar mechanism for updating agents' positions; however, FDO does it in a very different way, and it is statistically proven in this paper that FDO outperforms PSO, DA, GA, WOA, and SSA in many benchmark test functions and has comparative results on others.

The remainder of the paper is organized as follows. It begins by explaining the motivation behind the FDO algorithm and then debates the unique aspects that show the novelty of FDO algorithm. Then, the bee swarm features that inspired the FDO algorithm are presented. After that, the FDO algorithm is introduced by showing the pseudocode, equations, and rules. Furthermore, FDO for the single objective problem is described. Moreover, in the results and discussion section, detailed information is provided about FDO performance against other algorithms. In addition, FDO is applied to two real-world case scenarios. Finally, the main points about FDO, its limitation, and future works are described.

## II. BEE SWARMING

Since ancient times, this remarkable social insect has been one of the most famous creatures on the planet. Honeybees have been the subject of scientific observations. Likewise, considerable research and many books have been published about them, for example, "Behavior and the Social Life of Honeybees" by Ribbands in 1953. Snodgrass wrote "Anatomy of the Honey Bee in 1956, also "the wisdom of hive" by Thomas D. Seeley was written in 1995, and many other great works. Figure (1) shows the anatomy of a bee. Presenting the colony structure of bees and their biological details are beyond the scope of this paper. However, the swarming behavior of the bee life cycle will be discussed shortly since it is related to FDO.

As widely known, bees live and work in groups inside a colony called a hive (nest site). In brief, there are several types of bees: queen bees, worker bees, and scout bees. As their names suggest, the queen bee is responsible for making decisions and producing the next generation of bees. Worker bees work under the command of queen bees; they also create queen cells throughout the year. Finally, scout bees explore the environment and exploit the preferable targets, which is the most important feature of this work. Usually, when the number of bees in the hive increases and the inside colony conditions and outside weather conditions are suitable, then the queen lays eggs into the queen cells, and the bee colony starts the reproductive processes by swarming [25] [26].

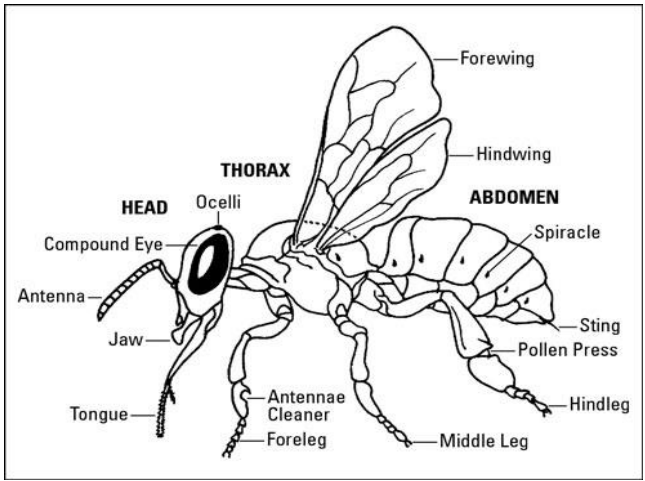

FIGURE. 1. Honey Bee Anatomy [27]

Swarming is mostly a late spring phenomenon; it is the process by which a new honeybee colony is formed. The queen bee leaves the old colony with a group of worker bees and some scout bees; Figure (2) shows the swarming cycle. A swarm typically consists of thousands to tens of thousands of bees. They settle 20–30 meters away from the natal hive temporarily for a few hours to a few days. They may gather in a tree or on a branch where they cluster around their queen, and then, they send 20 -50 scout bees out to find suitable new hives, usually after several tries, which might take several hours or up to three days. Eventually, with the guidance from the scouts, the rest of the bees flying overhead in the proper direction.

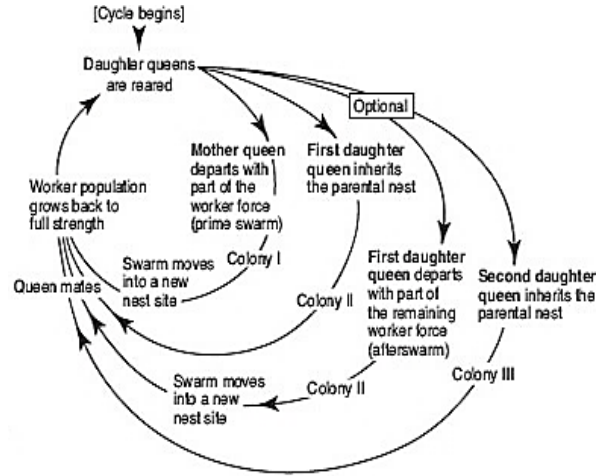

FIGURE. 2. Bee Swarming process cycle [28]

A swarm may fly a kilometer or more to the scouted location. Through direct observation, it can be said that the scout bee has several criteria for a suitable hive. For instance, a suitable hive has to be large enough to accommodate the whole swarm (minimum of 15 liters, preferably 40 liters in volume). It should have a small entrance (approximately 12.5 cm2), as well as being located at the lowest point of the hive, and obtain a certain amount of warmth from sunlight [26], [29].

What inspired us were the scouts' collective decision-making processes. When a number of scout bees discover some suitable hives, they will choose the most suitable hive, and they keep the swarm intact. Typically, scout bees communicate through moving their legs and wings, which is known as a bee dance. Usually, a decision will be made when approximately 80% of the scouts have agreed upon a certain hive location or when there is a quorum of 20-30 scout bees present at a potential hive [26], [29].

Algorithmically speaking, each hive that a scout bee exploits, represents a possible solution exploited by an artificial search agent, and the best hive represents the global optimum solution, as shown in Table (1). The hive specifications, such as its volume, entrance size, entrance location, and amount of sunlight, can also be considered as fitness functions of the solution. The scout's collective decision-making process, represented by fitness weight ($fw$) in the algorithm, $fw$ is discussed further in the next section.

TABLE 1
FDO-RELATED BEE BIOLOGICAL ENTITIES

| Nature | Algorithm |
| --- | --- |
| Scout bee | Search agent |
| Hive | Solution discovered |
| Hive specification | Fitness function |
| Scout collective decision | Fitness Weight |
| Selected hive | Optimum Solution |

## III. FITNESS DEPENDENT OPTIMIZER ALGORITHM

This algorithm replicates what a swarm of bees is doing during reproduction. The main part of this algorithm is taken from the process of scout bees searching for a new suitable hive among many potential hives. Every scout bee that searches for new hives represents a potential solution in this algorithm; furthermore, selecting the best hive among several good hives is considered as converging to the optimality.

The algorithm begins by randomly initializing an artificial scout population in the search space $X_i (i = 1, 2, \ldots n)$; each scout bee position represents a newly discovered hive (solution). Scout bees try to find better hives by randomly searching more positions; each time a better hive is found, the previously discovered hive is ignored; thus, each time the algorithm identifies a new, better solution, then the previously discovered solution will be ignored. In addition, if the current move is not leading the artificial scout bee to a better solution (hive), it will continue in its previous direction, hoping that the previous direction takes the scout to a better solution. However, if the previous direction does not lead to a better solution, it will then continue to the current solution, which is the best solution that has been found to that point.

In nature, scout bees search for hives randomly. In this algorithm, artificial scouts initially search the landscape randomly using a combination of a random walk and fitness weight mechanism. Accordingly, every time an artificial scout bee moves by adding pace to the current position, the scout hopes to explore a better solution. The movement of artificial scout bees is expressed as follows:

$$X_{i,t+1} = X_{i,t} + pace \quad (1)$$

Where $i$ represents the current search agent, $t$ represents the current iteration, $x$ represents an artificial scout bee (search agent), and $pace$, is the movement rate and direction of the artificial scout bee. $pace$ is mostly dependent on the fitness weight $fw$. However, the direction of $pace$ is completely dependent on a random mechanism. Thus, the $fw$ for minimization problems can be calculated as:

$$fw = \left| \frac{x^*_{i,t\ fitness}}{x_{i,t\ fitnees}} \right| - wf \quad (2)$$

The $x^*_{i,t\ fitness}$, is a fitness function value of the best global solution that has been discovered thus far. $x_{i,t\ fitness}$ is a value of the fitness function of the current solution, $wf$ is a weight factor, and its value is either 0 or 1, which is used for controlling the $fw$. If it is equal to 1, then it represents a high level of convergence and a low chance of coverage. Nonetheless, if $wf = 0$, then it is not affecting the Equation (2), thus it can be neglected, setting $wf = 0$ provides a more stable search. However, this is not always the case; sometimes, the opposite occurs because the fitness function value is completely optimization problem dependent. Nevertheless, the $fw$ value should be in the [0, 1] range; however, there are some cases where $fw = 1$, for example, when the current solution is the global best, or when the current and global best solutions are identical or have the same fitness value. Additionally, there is a chance that $fw = 0$, which occurs when $x^*_{i,t\ fitness} = 0$. Finally, division by zero should be avoided when $x_{i,t\ fitness} = 0$. Therefore, the following rules should be used:

$$\begin{cases} fw = 1 \text{ or } fw = 0 \text{ or } x_{i,t\ fitness} = 0, \quad pace = x_{i,t} * r \quad (3) \\ fw > 0 \text{ and } fw < 1 \begin{cases} r < 0, pace = (x_{i,t} - x^*_{i,t}) * fw * -1 \quad (4) \\ r \geq 0, \quad pace = (x_{i,t} - x^*_{i,t}) * fw \quad (5) \end{cases} \end{cases}$$

Here, $r$ is a random number in the [-1, 1] range. There are different implementations of the random walk; however, Levy flight has been chosen because it provides more stable movements because of its good distribution curve [3].

Regarding FDO mathematical complexity: For each iteration, it has an $O(p*n + p*CF)$ time complexity, where $p$ is the population size, $n$ is the dimension of the problem, and $CF$ is the cost of the objective function. Whereas, for all iterations, it has an $O(p*CF + p*pace)$ space complexity, where the $pace$ is the best previous paces stored. From here, FDO time complexity is proportional to the number of iterations. However, its space complexity will be the same during the course of iterations.

FDO has a simple calculation mechanism in terms of objective value calculations, it has only (fitness weight and one random number) to be calculated for each agent, whereas, in PSO for calculating each solution, there are global best, agent best, search factors $C1$ and $C2$, and random numbers ($R1$ and $R2$ parameters) to be calculated [8]. Also, in the DA, there are five different parameter weights to be calculated (separation, alignment, cohesion, attraction, distraction, and some random values), and most of these parameters have accumulative nature (summation and multiplication), and their values depend on all other agents' value, resulting in even more complex calculations [16].

## IV. FDO WITH SINGLE OBJECTIVE OPTIMIZATION PROBLEMS

The FDO with single objective optimization problems (FDOSOOP) begins by initializing artificial scouts at random locations on the search landscape, using upper and lower boundaries. For every iteration, the global best solution is selected; then, for every artificial scout bee, the $fw$ is calculated according to Equation (2). After that, the $fw$ value is checked to determine if $fw = 1$ or 0, also whether $x_{i,t\ fitness} = 0$. Then Equation (3) is used for generating the pace. However, if $fw > 0$ and $fw < 1$, then a random number $r$ will be generated in the [-1, 1] range. If $r < 0$ then Equation (4) is used to calculate the pace, in this case, $fw$ gets a negative sign, but if $r \geq 0$ then Equation (5) is used to calculate the pace, accordingly, $fw$ gets a positive sign. Randomly selecting negative or positive sign for a a $fw$ will guarantee that the artificial bee will search randomly in every direction.

In FDO randomization mechanism controls the pace size and direction, whereas in most cases, the randomization mechanism only controls the pace direction; in these cases, the pace size depends on the $fw$. Moreover, each time the artificial scout bee finds a new solution, it checks whether the new solution is better than the current solution, depending on the fitness function as shown in the pseudocode of the single objective FDO (see Figure (3)). If the new solution is better, then it is accepted, and the old solution is ignored. Additionally, one of the special features of FDO is that if the new solution is not better, then the artificial scout bee continues using the previous direction (using the previous $pace$ value if available), but only if it takes the scout bee to a better solution. In addition, if using the previous *pace* is not leading the scout bee to a better solution, then FDO maintains the current solution until the next iteration. In this algorithm, every time the solution is accepted, its *pace* value is saved for potential reuse in the next iteration.

When implementing FDO for maximization problems, two minor changes are needed. First, Equation (2) must be replaced by Equation (6), as Equation (6) is simply an inverse version of Equation (2).

$$fw = \left| \frac{x_{i,t\ fitness}}{x^*_{i,t\ fitness}} \right| - wf \qquad (6)$$

Second, the condition for selecting a better solution should be changed. The line: "if $(X_{t+1,i}\ fitness < X_{t,i}\ fitness)$" must be replaced with the line: "if $(X_{t+1,i}\ fitness > X_{t,i}\ fitness)$" in both occurrences in the pseudocode shown in Figure (3).

```
Initialize scout bee population X_{t,i}  (i = 1, 2, …, n)
while iteration (t) limit not reached
    for each artificial scout bee X_{t,i}
        find best artificial scout bee x*_{t,i}
        generate random-walk r in [-1, 1] range
        if( X_{t,i} fitness == 0) (avoid divide by zero)
            fitness weight = 0
        else
            calculate fitness weight. equation (2)
        end if
        if (fitness weight = 1 or fitness weight = 0)
            calculate pace using equation (3)
        else
            if (random number >= 0)
                calculate pace using equation (5)
            else
                calculate pace using equation (4)
            end if
        end if
        calculate X_{t+1,i}   equation (1)
        if(X_{t+1,i} fitness < X_{t,i} fitness)
            move accepted and pace saved
        else
            calculate X_{t+1,i} equation (1) …
                    … with previous pace
            if (X_{t+1,i} fitness < X_{t,i} fitness)
                move accepted and pace  saved
            else
                maintain current position (don't move)
            end if
        end if
    end for
end while
```

**FIGURE. 3. Pseudocode of FDOSOOP**

## V. RESULTS AND DISCUSSION

To test the performance of this algorithm, a number of standard benchmark test functions exist in the literature is used. Additionally, our results are compared to five other well-known algorithms in the literature: PSO, GA, DA, WOA, and SSA. It is worth mentioning that results of (19 classical benchmark test functions) PSO, GA, and DA is taken from this work [16], however, we conducted the CEC-C06 tests. Also, all test results were compared using the Wilcoxon rank-sum test to prove their statistical significance. Moreover, four measurement metrics were used for further observation. Finally, the FDO was used for optimizing two real-world applications; thus, the section consists of five parts as follows.

*1) Classical Benchmark Test Functions*

Three sets of test functions are selected to test the performance of the FDO algorithm [16]. The test functions have different characteristics, for instance, unimodal test functions, multimodal test functions, and composite test functions. Each set of these test functions is used to benchmark certain perspectives of the algorithm. Unimodal benchmark functions, for example, are used for testing the exploitation level and convergence of the algorithm, as their name might imply that they have a single optimum . However, multimodal benchmark functions have multi optimal solutions, and they are used for testing the local optima avoidance and exploration levels. As in multimodal algorithms, there are many optimum solutions; one of them is a global optimum solution and most local optimum solutions. An algorithm must avoid local optimum solutions and converge to a global optimum solution. Furthermore, the composite benchmark functions are mostly combined, shifted, rotated, and biased versions of other test functions. Composite benchmark functions provide diverse shapes for different regions of the search landscape; they also have a very large number of local optima. This type of benchmark function demonstrates that complications exist in real-world search spaces (see Table (6, 7 and 8) in the appendix [16]).

Each algorithm in Table (2) has been tested 30 times by using 30 search agents each with 10 dimensions; in each test, the algorithm was allowed to look for the best optimum solution in 500 iterations, and then, the average and standard deviation were calculated. Regarding parameter sets, GA, PSO, and DA parameter sets described in this paper [16]. But for FDO parameters, there is only $wf$ to be tuned. In Table (2), for all test functions $wf$ was equal to 0 except test function (2 and 6) where $wf$ equal to 1. Every test function was minimized towards 0.0 except TF8, which was minimized towards -418.9829 (see Appendix Tables 6, 7 and 8 for more details about the test function conditions). For example, some test functions were shifted by some degrees from the origin point to prove that the algorithms were not biased towards the origin.

In Table (2), the results of FDO, DA, PSO, and GA are presented. The TF1 to TF6 results showed that FDO generally provided better results than the other algorithms; however, the TF7 results showed the other algorithms were better. FDO in TF8 showed poor performance even though it had better results than PSO. In contrast, TF9 FDO provided a better result than both GA and DA, and comparative results were produced by PSO. In TF10 to TF13 and TF18, FDO provided relatively comparative results to the other algorithms. However, the results of TF14 to TF17 and TF19 confirm that the FDO algorithm outperformed DA, PSO, and GA in all cases.

TABLE 2
CLASSICAL BENCHMARK RESULTS OF SELECTED ALGORITHMS WITH FDO [16]

| Test Function | FDO Ave | FDO Std | DA Ave | DA Std | PSO Ave | PSO Std | GA Ave | GA Std |
|---|---|---|---|---|---|---|---|---|
| TF1 | 7.47E-21 | 7.26E-19 | 2.85E-18 | 7.16E-18 | 4.2E-18 | 1.31E-17 | 748.5972 | 324.9262 |
| TF2 | 9.388E-6 | 6.90696E-6 | 1.49E-05 | 3.76E-05 | 0.003154 | 0.009811 | 5.971358 | 1.533102 |
| TF3 | 8.5522E-7 | 4.39552E-6 | 1.29E-06 | 2.1E-06 | 0.001891 | 0.003311 | 1949.003 | 994.2733 |
| TF4 | 6.688E-4 | 0.0024887 | 0.000988 | 0.002776 | 0.001748 | 0.002515 | 21.16304 | 2.605406 |
| TF5 | 23.50100 | 59.7883701 | 7.600558 | 6.786473 | 63.45331 | 80.12726 | 133307.1 | 85,007.62 |
| TF6 | 1.422E-18 | 4.7460E-18 | 4.17E-16 | 1.32E-15 | 4.36E-17 | 1.38E-16 | 563.8889 | 229.6997 |
| TF7 | 0.544401 | 0.3151575 | 0.010293 | 0.004691 | 0.005973 | 0.003583 | 0.166872 | 0.072571 |
| TF8 | -2285207 | 206684.91 | -2857.58 | 383.6466 | -7.1E+11 | 1.2E+12 | -3407.25 | 164.4776 |
| TF9 | 14.56544 | 5.202232 | 16.01883 | 9.479113 | 10.44724 | 7.879807 | 25.51886 | 6.66936 |
| TF10 | 3.996E-15 | 6.3773E-16 | 0.23103 | 0.487053 | 0.280137 | 0.601817 | 9.498785 | 1.271393 |
| TF11 | 0.568776 | 0.1042672 | 0.193354 | 0.073495 | 0.083463 | 0.035067 | 7.719959 | 3.62607 |
| TF12 | 19.83835 | 26.374228 | 0.031101 | 0.098349 | 8.57E-11 | 2.71E-10 | 1858.502 | 5820.215 |
| TF13 | 10.2783 | 7.42028 | 0.002197 | 0.004633 | 0.002197 | 0.004633 | 68,047.23 | 87,736.76 |
| TF14 | 3.7870E-7 | 6.3193E-7 | 103.742 | 91.24364 | 150 | 135.4006 | 130.0991 | 21.32037 |
| TF15 | 0.001502 | 0.0012431 | 193.0171 | 80.6332 | 188.1951 | 157.2834 | 116.0554 | 19.19351 |
| TF16 | 0.006375 | 0.0105688 | 458.2962 | 165.3724 | 263.0948 | 187.1352 | 383.9184 | 36.60532 |
| TF17 | 23.82013 | 0.2149425 | 596.6629 | 171.0631 | 466.5429 | 180.9493 | 503.0485 | 35.79406 |
| TF18 | 222.9682 | 9.9625E-6 | 229.9515 | 184.6095 | 136.1759 | 160.0187 | 118.438 | 51.00183 |
| TF19 | 22.7801 | 0.0103584 | 679.588 | 199.4014 | 741.6341 | 206.7296 | 544.1018 | 13.30161 |

*2) CEC-C06 2019 Benchmark Test Functions*

A group of 10 modern CEC benchmark test functions is used as an extra evaluation on FDO, these test functions were improved by professor Suganthan and his colleges for a single objective optimization problem [1], the test functions are known as "The 100-Digit Challenge", which are intended to be used in annual optimization competition. See Table (9) in the appendix.

Functions CEC04 to CEC10 are shifted and rotated, whereas functions CEC01 to CEC03 are not. However, all test functions are scalable. The parameter set where defined by the CEC benchmark developer, as functions CEC04 to CEC10 where set as 10-dimensional minimization problem in [-100, 100] boundary range, however, CEC01 to CEC03 have different dimensions as shown in the Appendix in Table 9. For more convenient, all CEC global optimum where unified toward point 1. FDO is competed with three modern optimization algorithms: DA, WOA, and SSA. The reasons behind selecting these algorithms are: 1) They are all PSO-based algorithms same as FDO. 2) All of them are well cited in the literature. 3) They are Proven to have an outstanding performance both on benchmark test functions and real-world problems. 4) These algorithm implementations are publicly provided by their authors. Regarding algorithms parameter settings, their default parameter settings were not modified during the tests, all competitors are set the same as the settings used in their original papers [16] [17] [18]. Interested readers can find these algorithms MATLAB implementations and their parameter setting specification here [30]. Additionally, FDO default parameter set $wf = 0$ is used for all test functions.

TABLE 3
IEEE CEC 2019 BENCHMARK RESULTS

| Test Function | FDO Average | FDO STD | DA Average | DA STD | WOA Average | WOA STD | SSA Average | SSA STD |
|---|---|---|---|---|---|---|---|---|
| CEC01 | 4585.27 | 20707.627 | 543×10[8] | 669×10[8] | 411×10[8] | 542×10[8] | 605×10[7] | 475×10[7] |
| CEC02 | 4.0 | 3.22414E-9 | 78.0368 | 87.7888 | 17.3495 | 0.0045 | 18.3434 | 0.0005 |
| CEC03 | 13.7024 | 1.6490E-11 | 13.7026 | 0.0007 | 13.7024 | 0.0 | 13.7025 | 0.0003 |
| CEC04 | 34.0837 | 16.528865 | 344.3561 | 414.0982 | 394.6754 | 248.5627 | 41.6936 | 22.2191 |
| CEC05 | 2.13924 | 0.085751 | 2.5572 | 0.3245 | 2.7342 | 0.2917 | 2.2084 | 0.1064 |
| CEC06 | 12.1332 | 0.600237 | 9.8955 | 1.6404 | 10.7085 | 1.0325 | 6.0798 | 1.4873 |
| CEC07 | 120.4858 | 13.59369 | 578.9531 | 329.3983 | 490.6843 | 194.8318 | 410.3964 | 290.5562 |
| CEC08 | 6.1021 | 0.756997 | 6.8734 | 0.5015 | 6.909 | 0.4269 | 6.3723 | 0.5862 |
| CEC09 | 2.0 | 1.5916E-10 | 6.0467 | 2.871 | 5.9371 | 1.6566 | 3.6704 | 0.2362 |
| CEC10 | 2.7182 | 8.8817E-16 | 21.2604 | 0.1715 | 21.2761 | 0.1111 | 21.04 | 0.078 |

Each algorithm where allowed to search the landscape for 500 iterations using 30 agents. As shown in Table (3), FDO outperforms other algorithms except in CEC06. Even though other algorithms have a comparative result in CEC03, CEC05, and CEC09 benchmarks, for example, WOA has the same result as FDO in CEC03, but the WOA standard deviation is equal to 0, this shows that WOA has the same result every time it uses with no chance for further improvements.

*3) Statistical Tests*

To show that the results presented in Table (2) and Table (3) are statistically significant, the p values of the Wilcoxon rank-sum test are found for all test functions, and the results of a statistical comparison are shown in Table (4) and Table (5). In Table (4), the comparison is conducted only between the FDO and DA algorithms because the DA algorithm was already tested against both PSO and GA in this paper [16]. According to the mentioned work, it has been proven that the DA results are statistically significant compared with PSO and GA.

TABLE 4
THE WILCOXON RANK-SUM TEST FOR CLASSICAL BENCHMARKS

| Function | DFO vs. DA (P value) |
| --- | --- |
| TF1 | 0.000513631 |
| TF2 | 0.7111046 |
| TF3 | 8.2371E-05 |
| TF4 | 1.0750E-08 |
| TF5 | 0.0023817 |
| TF6 | 1.7620E-04 |
| TF7 | 6.6643E-12 |
| TF8 | 1.4805E-34 |
| TF9 | 5.8820E-08 |
| TF10 | 5.6002E-17 |
| TF11 | 0.012793 |
| TF12 | 9.4381E-05 |
| TF13 | 1.0426E-37 |
| TF14 | 1.1142E-21 |
| TF15 | 0.000094477 |
| TF16 | 0.0000E+00 |
| TF17 | 1.0668E-120 |
| TF18 | 0.0000E+00 |
| TF19 | 5.5373E-166 |

Again, as shown in Table (4), the FDO results are considered significant in all statistical tests (unimodal, multimodal and composite test functions), except in TF2, that is because the results are more than 0.05. There are two unusual results in the composite test functions in both TF16 and TF18 because the DA algorithm provided the same fitness function value for each of the 30 different individual tests.

Table (5) shows the Wilcoxon rank-sum test of FDO against DA, WOA, and SSA for 10 CEC benchmark test functions, the results show that FDO performances are statistically significant in all cases, except in test function CEC03 for DA and WOA algorithms, and test function CEC04 and CEC08 for WOA algorithm. The results of Table (4) and Table (5) prove that FDO results are statistically significant, consequently, the existence of the FDO algorithm is statistically feasible.

TABLE 5
THE WILCOXON RANK-SUM TEST (P-VALUE) FOR CEC 2019

| Func. | FDO Vs. DA | FDO Vs. WOA | FDO Vs. SSA |
| --- | --- | --- | --- |
| CEC01 | 4.03455E-05 | 0.000108312 | 3.1831E-09 |
| CEC02 | 1.81428E-05 | 3.17E-252 | 1.1E-196 |
| CEC03 | 0.244847 | 0.363647 | 1E-306 |
| CEC04 | 0.000124 | 0.095911 | 7.4365E-11 |
| CEC05 | 6.05468E-09 | 0.007884 | 2.3682E-15 |
| CEC06 | 2.90314E-09 | 2.196E-28 | 1.9679E-08 |
| CEC07 | 2.69474E-10 | 1.0424E-06 | 6.2691E-15 |
| CEC08 | 2.3638E-05 | 0.131704 | 5.2331E-06 |
| CEC09 | 1.80487E-10 | 3.7992E-43 | 7.4029E-19 |
| CEC10 | 2.2248E-111 | 6.397E-131 | 2.459E-122 |

*4) Quantitative Measurement Metrics*

For more detailed analyses and in-depth observation of the FDO algorithm, four more quantitative metrics were used, as shown in Figures (4, 5, 6 and 7). In each experiment, the first test function is selected from the unimodal benchmark functions (FT1 to FT7), the second test function selected is from the multimodal test functions (TF8 to TF13), and the last test function is selected from the composite benchmark functions (TF14 to TF19). The experiment was conducted using 10 search agents, each allowed to search the two-dimensional landscape through 150 iterations.

The first metric measures the convergence and illustrates how well the artificial scout covers the search landscape. This is merely a search history of artificial scout movements because the position of the artificial scouts is recorded from the beginning to the end of the test. As presented in Figure (4), the scout quickly explores the overall area first and then gradually moves towards optimality.

The second metric measures the value of the search agent (fitness function value), as shown in Figure (5). The values start with large values and then steadily decrease. This behavior guarantees that FDO will eventually reach optimality [31].

The third test metric is shown in Figure (6) and shows that the average fitness value of all FDO agents decreased dramatically over the course of the iterations, which verifies that the algorithm not only improves the global best agent ($x_i^*$) but also improves the overall agent fitness values.

The fourth metric measures the convergence of the global best agent through the course of the iteration. This proves that $x_i^*$ becomes more accurate as the number of iterations increases again, clear abrupt changes can be seen due to the emphasis on the local search and exploitation, see Figure (7).

Overall, in this section, measurement metrics showed that FDO is capable of effectively exploring the search space, improving the overall solution, avoiding local optimum and fairly converging towards optimality.

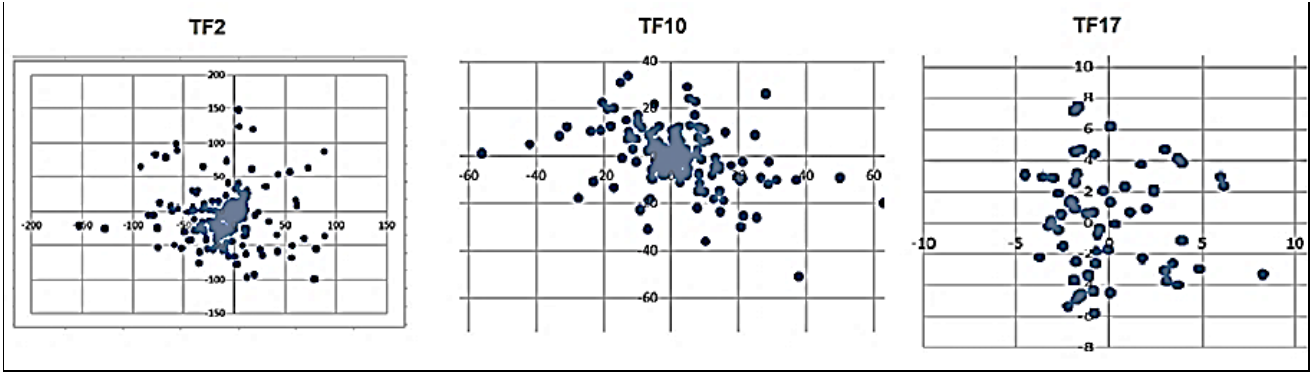

**FIGURE. 4.** Search history of the FDO algorithms on unimodal, multimodal, and composite test functions

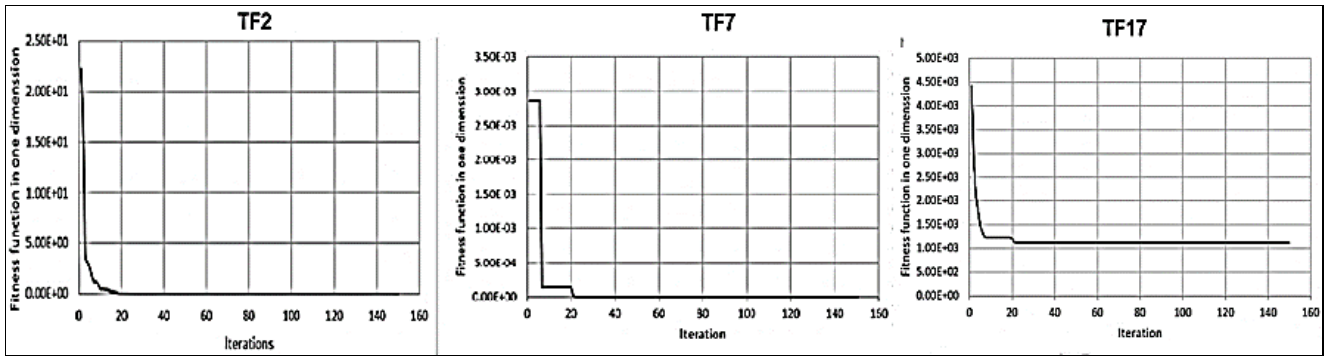

**FIGURE. 5.** The trajectory of FDO's search agents on unimodal, multimodal, and composite test functions

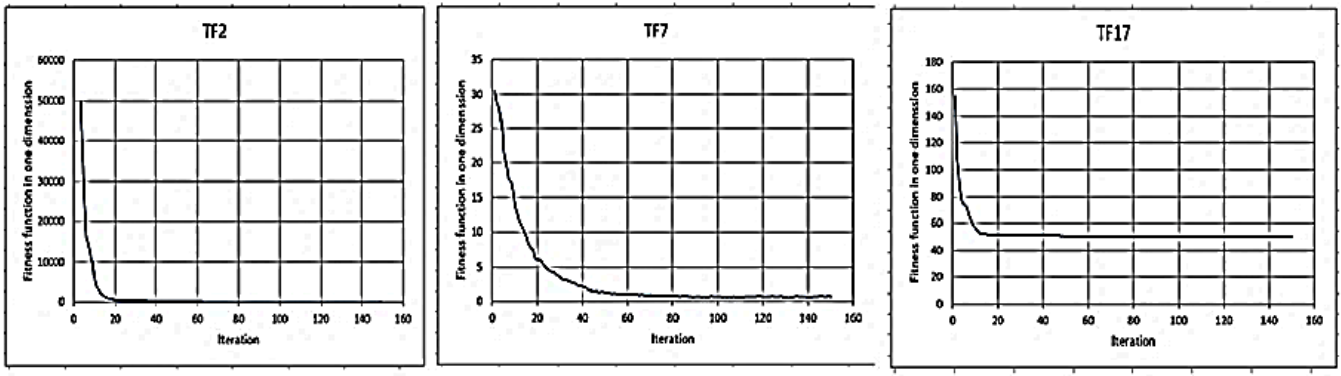

**FIGURE. 6.** Average fitness of FDO's search agents on unimodal, multimodal, and composite test functions

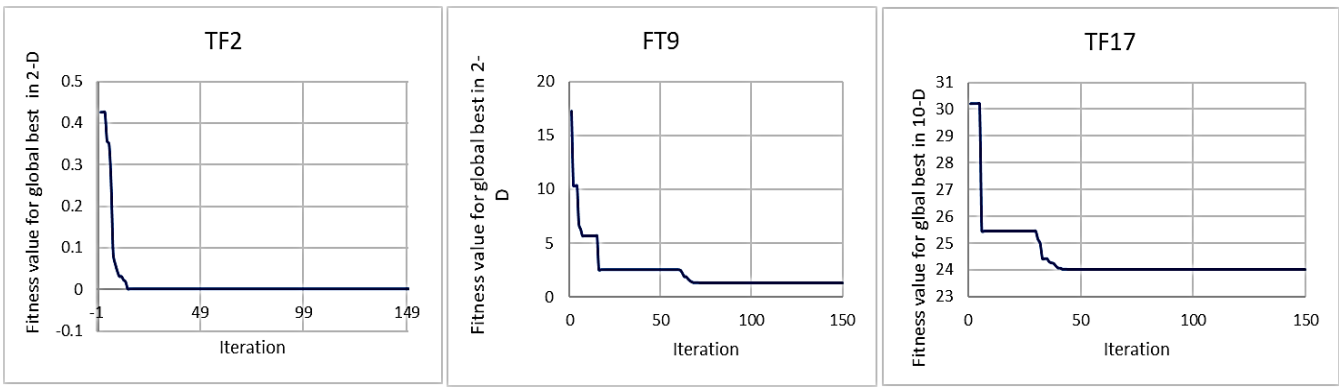

**FIGURE. 7.** *C*onvergence curve of the FDO's algorithms on unimodal, multimodal, and composite test functions

## 5) *FDO Real World Application*

Similar to any other metaheuristic algorithm, FDO can be used to solve real-world application problems. In this section, FDO is applied to two different applications:

### A- FDO USAGE ON APERIODIC ANTENNA ARRAY DESIGNS.

Since the 1960s, with the advances in both radar techniques and radio astronomy, aperiodic antenna arrays have received considerable attention, as shown in Figure (8); there are two types of aperiodic antenna arrays: nonuniform antenna arrays and thin antenna arrays.

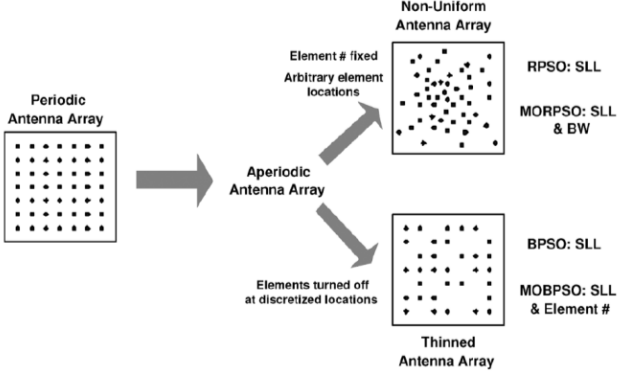

**FIGURE. 8. Nonuniform antenna array and a thinned antenna array [22].**

In particular, to obtain the peak sidelobe level (SLL) in nonuniform arrays, the element position should be optimized in terms of a real number vector, as shown in Figure (9). Moreover, to avoid grating lobes, a certain element spacing limit exists for conventional periodic arrays (see constraints in Equation (7)). Interested readers can review [32] for more details on this problem.

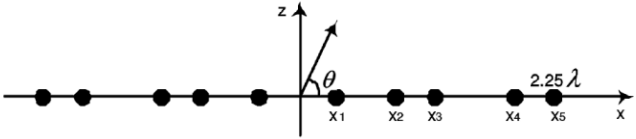

**FIGURE. 9. Array configurations for 10-elements [32].**

Again, as shown in Figure (9), there are 10 elements of a nonuniform isotropic array, and only four element locations need to be optimized on each side. Since the outer-most element is fixed at location $2.25\lambda_0$ with an average element spacing of $d_{avg} = 0.5\lambda_0$, this is a four-dimensional optimization problem with the following constraints:

$$x_i \in (0, 2.25) | x_i - x_j | > 0.25\lambda_0 \quad (7)$$
$$min\{x_i\} > 0.125\lambda_0. \ i = 1,2,3,4. i \neq j.$$

The constraints show that there is a boundary between 0 and 2.25 for every element. However, each element cannot be smaller than $0.125\lambda_0$ or larger than $2.0\lambda_0$; that is because of $2.25\lambda_0$ is a fixed element and two adjacent elements cannot get closer than $0.25\lambda_0$. The fitness function problem is described as:

$$f = max\{20 \log|AF(\theta)|\} \quad (8)$$

where

$$AF(\theta) = \sum_{i=1}^{4} \cos[2\pi x_i(\cos\theta - \cos\theta_s)] \quad (9)$$
$$+ \cos[2.25 \times 2\pi(\cos\theta - \cos\theta_s)]$$

Consider that $\theta_s = 90°$ in this work is defined in Figure (9) [32].

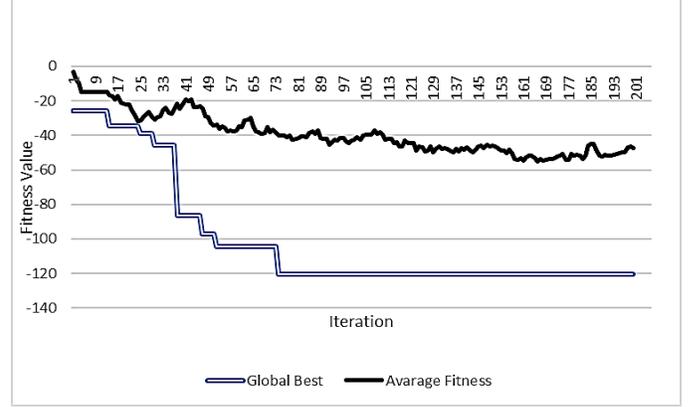

**FIGURE. 10. Global best with average fitness results for 200 Iteration with 20 artificial scout bees on aperiodic antenna array designs.**

The DFO algorithm is used to optimize this problem, considering the constraints mentioned in Equation (7). Twenty artificial scout search agents are used for 200 iterations, and the presented result in Figure (10) includes the global best fitness in each iteration and the average fitness value according to Equation (8). The result shows that the global best solution reached its optimum solution in iteration 78 with element positions = { 0.713, 1.595, 0.433, 0.130}.

### B- FDO ON FREQUENCY MODULATED SOUND WAVES

FDO is used on frequency-modulated sound waves (FM) to optimize the parameter of an FM synthesizer, which has an essential role in several modern music systems; this problem has six parameters to be optimized as indicated in Equation (10).

$$X = \{a_1, w_1, a_2, w_2, a_3, w_3\} \quad (10)$$

The objective of this problem is to generate a sound, as in Equation (11), that is similar to the target sound, as in Equation (12).

$$y(t) = a_1.\sin(w_1.t. + a_2.\sin(w_2.t.\theta \quad (11)$$
$$+ a_3.\sin(w_3.t.\theta)))$$

$$y_o(t) = (1.0).\sin((5.0).t. + (1.5).\sin((4.8).t.\theta \quad (12)$$
$$+ (2.0).\sin((4.9).t.\theta)))$$

where the parameters should be in the range $[-6.4, 6.35]$ and $\theta = 2\pi/100$, the fitness function can be calculated using Equation (13), which is simply the summation of the square root between the result of Equation (11) and Equation (12), while t = 100 turns.

$$f(\vec{x}) = \sum_{t=0}^{100}(y(t) - y_o(t))^2 \quad (13)$$

Interested readers can find more details on this problem in [33].

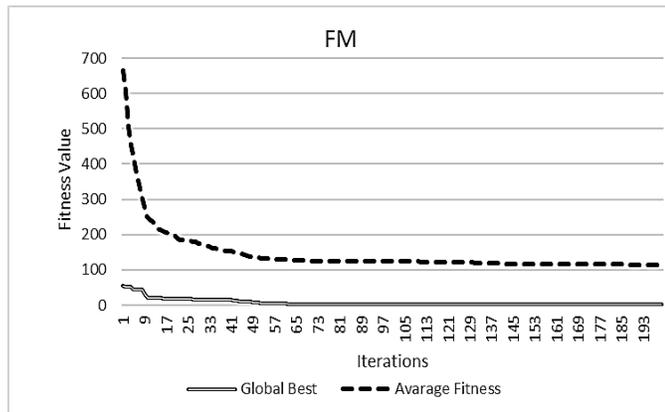

**FIGURE. 11.** Global best with average fitness results for 200 Iteration with 30 artificial scout bees on the FM synthesis problem.

FDO is applied to the problem with 30 agents for 200 iterations, and records of the global best solutions and average fineness values can be seen in Figure (11). Parameter-set $X = \{a_1 = 0.974, w_1 = -0.241, a_2 = -4.3160, w_2 = -0.0193, a_3 = -0.5701, w_3 = 4.937\}$ were also generated at iteration 200. The global best value converges to the near-global optimal value from iteration 64.

## VI. CONCLUSION

A new swarm intelligent algorithm was proposed called the fitness dependent optimizer; it is inspired by the bee reproductive swarming process, where scout bees search for a new nest site. Additionally, the algorithm is inspired by their collective decision-making. It has no algorithmic connection with the ABC algorithm. FDO employs fitness function values to generate weights that drive the search agents towards optimality. Additionally, FDO depends on the randomization mechanism in the initialization, exploration and exploitation phases. A group of 19 single objective benchmark testing functions was used to test the performance of the FDO. The benchmark testing functions were divided into three subgroups (unimodal, multimodal and composite test functions). Additionally, FDO tested on 10 modern CEC-C06 benchmarks. The FDO results compared to two well-known algorithms (PSO and GA) and three modern algorithms (DA, WOA, and SSA), FDO outperformed the competing algorithms in the majority of cases and produced a comparative result on the others. The test results were compared using the Wilcoxon rank-sum test to prove their statistical significance. Four additional experiments were conducted on the FDO algorithm to measure, prove and verify the performance and credibility. Furthermore, FDO was practically applied to two real-world examples as evidence that the algorithm can address real-life applications.

Generally, we found that the number of search agents was related somehow to FDO performance after testing on many standard test functions and real-world applications. Thus, using a small number of agents (below five) would notably decrease the accuracy of the algorithm, and a large number of search agents would improve the accuracy and cost more time and space, partially because the algorithm depends on the fitness weight on the significant part of its searching mechanism; in view of this, it is known as the fitness dependent optimizer.

Future works will adapt, implement and test both multi-objective and binary objective optimization problems on FDO. Finally, integrating evolutionary operators into FDO and hybridizing it with other algorithms can be considered as potential future research.

# VIII. APPENDIX

TABLE 6
UNIMODAL BENCHMARK FUNCTIONS [16]

| Functions | Dimension | Range | Shift position | $f_{min}$ |
|---|---|---|---|---|
| $TF1(x) = \sum_{i=1}^{n} x_i^2$ | 10 | [-100, 100] | [-30, -30, … -30] | 0 |
| $TF2(x) = \sum_{i=1}^{n} \|x_i\| + \prod_{i=1}^{n} \|x_i\|$ | 10 | [-10,10] | [-3, -3, … -3] | 0 |
| $TF3(x) = \sum_{i=1}^{n} \left( \sum_{j-1}^{i} x_j \right)^2$ | 10 | [-100, 100] | [-30, -30, … -30] | 0 |
| $TF4(x) = \max_i \{\|x\|, 1 \leq i \leq n\}$ | 10 | [-100, 100] | [-30, -30, … -30] | 0 |
| $TF5(x) = \sum_{i=1}^{n-1} [100(x_{i+1} - x_1^2)^2 + (x_i - 1)^2]$ | 10 | [-30,30] | [-15, -15, … -15] | 0 |
| $TF6(x) = \sum_{i=1}^{n} ([x_i + 0.5])^2$ | 10 | [-100, 100] | [-750, … -750] | 0 |
| $TF7(x) = \sum_{i=1}^{n} i x_i^4 + \text{random}[0,1]$ | 10 | [-1.28,1.28] | [-0.25, …-0.25] | 0 |

TABLE 7
MULTIMODAL BENCHMARK FUNCTIONS (10 DIMENSIONAL) [16]

| Functions | Range | Shift position | $f_{min}$ |
|---|---|---|---|
| $TF8(x) = \sum_{i=1}^{n} -x_i^2 \sin\left(\sqrt{\|x_i\|}\right)$ | [-500, 500] | [-300, … -300] | -418.9829 |
| $TF9(x) = \sum_{i=1}^{n} [x_i^2 - 10\cos(2\pi x_i) + 10]$ | [-5.12,5.12] | [-2, -2, …-2] | 0 |
| $TF10(x) = -20 exp\left(-0.2 \sqrt{\sum_{i=1}^{n} x_i^2}\right) - exp\left(\frac{1}{n}\sum_{i=1}^{n} \cos(2\pi x_i)\right) + 20 + e$ | [-32, 32] | | 0 |
| $TF11(x) = \frac{1}{4000} \sum_{i=1}^{n} x_i^2 - \prod_{i=1}^{n} \cos\left(\frac{x_i}{\sqrt{i}}\right) + 1$ | [-600, 600] | [-400, … -400] | 0 |
| $TF12(x) = \frac{\pi}{n} \{10 \sin(\pi y_1) + \sum_{i=1}^{n-1}(y_i - 1)^2 [1 + 10 \sin^2(\pi y_{i+1})] + (y_n - 1)^2\} + \sum_{i=1}^{n} u(x_i, 10, 100, 4).$ $y_i = 1 + \frac{x+1}{4}.$ $u(x_i, a, k, m) = \begin{cases} k(x_i - a)^m & x_i > a \\ 0 & -a < x_i < a \\ k(-x_i - a)^m & x_i < -a \end{cases}$ | [-50,50] | [-30, 30, … 30] | 0 |
| $TF13(x) = 0.1\left\{\sin^2(3\pi x1) + \sum_{i=1}^{n}(x_i-1)^2[1+\sin^2(3\pi x_i+1)] + (x_n-1)^2[1+\sin^2(2\pi x_n)]\right\} + \sum_{i=1}^{n} u(x_i,5,100,4).$ | [-50,50] | [-100, … -100] | 0 |

TABLE 8
COMPOSITE BENCHMARK FUNCTIONS [16]

| **Functions** | **Dimension** | **Range** | $f_{min}$ |
|---|---|---|---|
| **TF14 (CF1)** <br> $f1, f2, f3 \dots f10 = $ Sphere function <br> $\delta1, \delta2, \delta3 \dots \delta10 = [1,1,1,\dots.1]$ <br> $\lambda1, \lambda2, \lambda3 \dots \lambda10 = \left[\frac{5}{100}, \frac{5}{100}, \frac{5}{100}, \dots \frac{5}{100}\right]$ | 10 | [-5, 5] | 0 |
| **TF15 (CF2)** <br> $f1, f2, f3 \dots f10 = $ Griewank's function <br> $\delta1, \delta2, \delta3 \dots \delta10 = [1,1,1,\dots.1]$ <br> $\lambda1, \lambda2, \lambda3 \dots \lambda10 = \left[\frac{5}{100}, \frac{5}{100}, \frac{5}{100}, \dots \frac{5}{100}\right]$ | 10 | [-5, 5] | 0 |
| **TF16 (CF3)** <br> $f1, f2, f3 \dots f10 = $ Griewank's function <br> $\delta1, \delta2, \delta3 \dots \delta10 = [1,1,1,\dots.1]$ <br> $\lambda1, \lambda2, \lambda3 \dots \lambda10 = [1,1,1,\dots.1]$ | 10 | [-5, 5] | 0 |
| **TF17 (CF4)** <br> $f1, f2 = $ Ackley's function <br> $f3, f4 = $ Rastrigin's function <br> $f5, f6 = $ Weierstrass function <br> $f7, f8 = $ Griewank's function <br> $f9, f10 = $ Sphere function <br> $\delta1, \delta2, \delta3 \dots \delta10 = [1,1,1,\dots.1]$ <br> $\lambda1, \lambda2, \lambda3 \dots \lambda10 = \left[\frac{5}{32}, \frac{5}{32}, 1, 1, \frac{5}{0.5}, \frac{5}{0.5}, \frac{5}{100}, \frac{5}{100}, \frac{5}{100}, \frac{5}{100}\right]$ | 10 | [-5, 5] | 0 |
| **TF18 (CF5)** <br> $f1, f2 = $ Rastrigin's function <br> $f3, f4 = $ Weierstrass function <br> $f5, f6 = $ Griewank's function <br> $f7, f8 = $ Ackley's function <br> $f9, f10 = $ Sphere function <br> $\delta1, \delta2, \delta3 \dots \delta10 = [1,1,1,\dots.1]$ <br> $\lambda1, \lambda2, \lambda3 \dots \lambda10 = \left[\frac{1}{5}, \frac{1}{5}, \frac{5}{0.5}, \frac{5}{0.5}, \frac{5}{100}, \frac{5}{100}, \frac{5}{32}, \frac{5}{32}, \frac{5}{100}, \frac{5}{100}\right]$ | 10 | [-5, 5] | 0 |
| **TF19 (CF6)** <br> $f1, f2 = $ Rastrigin's function <br> $f3, f4 = $ Weierstrass function <br> $f5, f6 = $ Griewank's function <br> $f7, f8 = $ Ackley's function <br> $f9, f10 = $ Sphere function <br> $\delta1, \delta2, \delta3 \dots \delta10 = [0.1, 0.2, 0.3, 0.4, 0.5, 0.6, 0.7, 0.8, 0.9, 1]$ <br> $\lambda1, \lambda2, \lambda3 \dots \lambda10 = \left[0.1*\frac{1}{5}, 0.2*\frac{1}{5}, 0.3*\frac{5}{0.5}, 0.4*\frac{5}{0.5}, 0.5*\frac{5}{100}, 0.6 * \frac{5}{100}, 0.7*\frac{5}{32}, 0.8*\frac{5}{32}, 0.9*\frac{5}{100}, 1*5/100\right]$ | 10 | [-5, 5] | 0 |

TABLE 9
CEC-C06 2019 BENCHMARKS "THE 100-DIGIT CHALLENGE:" [1].

| No. | Functions | Dimension | Range | $f_{min}$ |
|---|---|---|---|---|
| 1 | STORN'S CHEBYSHEV POLYNOMIAL FITTING PROBLEM | 9 | [-8192, 8192] | 1 |
| 2 | INVERSE HILBERT MATRIX PROBLEM | 16 | [-16384, 16384] | 1 |
| 3 | LENNARD-JONES MINIMUM ENERGY CLUSTER | 18 | [-4,4] | 1 |
| 4 | RASTRIGIN'S FUNCTION | 10 | [-100, 100] | 1 |
| 5 | GRIEWANGK'S FUNCTION | 10 | [-100, 100] | 1 |
| 6 | WEIERSTRASS FUNCTION | 10 | [-100, 100] | 1 |
| 7 | MODIFIED SCHWEFEL'S FUNCTION | 10 | [-100, 100] | 1 |
| 8 | EXPANDED SCHAFFER'S F6 FUNCTION | 10 | [-100, 100] | 1 |
| 9 | HAPPY CAT FUNCTION | 10 | [-100, 100] | 1 |
| 10 | ACKLEY FUNCTION | 10 | [-100, 100] | 1 |

NOTE: Interested reader can see this technical paper [1] for more information about the CEC benchmarks.